\documentclass[runningheads]{llncs}
\usepackage[T1]{fontenc}
\usepackage{graphicx}
\usepackage{booktabs}
\usepackage[misc]{ifsym}
\newcommand{\corr}{(\Letter)}
\usepackage{hyperref}

\usepackage{hyperref}
\usepackage[table,xcdraw]{xcolor}
\definecolor{lightblue}{RGB}{173, 216, 230}
\definecolor{lightgreen}{RGB}{144, 238, 144}
\definecolor{lightpink}{RGB}{255, 182, 193}

\usepackage{multirow}
\usepackage{amssymb}

\usepackage{amsmath}

\usepackage[ruled,vlined]{algorithm2e} 
\usepackage{amsmath} 
\usepackage{amssymb} 

\usepackage{mwe}
\usepackage{soul}
\newcommand{\ot}[1]{\textcolor{black}{#1}}

\begin{document}

\title{XStacking: Explanation-Guided Stacked Ensemble Learning}

\titlerunning{XStacking: Explanation-Guided Stacked Ensemble Learning}

\author{Moncef Garouani\inst{1}\corr \and
Ayah Barhrhouj\inst{2} \and
Olivier Teste\inst{3}}

\authorrunning{Moncef Garouani, et al.}

\institute{IRIT, UMR 5505 CNRS, Université Toulouse Capitole, Toulouse 31000, France \email{moncef.garouani@irit.fr}
\and
Aix-Marseille University, CNRS, LIS UMR 7020, Marseille 13007, France 
\and
IRIT, UMR 5505 CNRS, Université de Toulouse, UT2J, Toulouse 31000, France
}

\maketitle              
\setcounter{footnote}{0}

\begin{abstract}

Ensemble Machine Learning (EML) techniques, especially stacking, have been shown to improve predictive performance by combining multiple base models. However, they are often criticized for their lack of interpretability. In our recent publication\footnote{Garouani, M., Barhrhouj, A., \& Teste, O. "XStacking: An effective and inherently explainable framework for stacked ensemble learning." Information Fusion, Volume 124, 2025, 103358. DOI: \href{https://doi.org/10.1016/j.inffus.2025.103358}{10.1016/j.inffus.2025.103358}} in \textit{Information Fusion}, we introduce  \textbf{XStacking}, an effective and inherently explainable framework that addresses this limitation by integrating dynamic feature transformation with model-agnostic Shapley additive explanations. This enables stacked models to retain their predictive accuracy while becoming inherently explainable. We demonstrate the effectiveness of the framework on 29 datasets, achieving improvements in both the predictive effectiveness of the learning space and the interpretability of the resulting models. XStacking offers a practical and scalable solution for responsible ML.


\keywords{Machine learning \and
Ensemble learning \and
Stacking \and
Explainable Artificial Intelligence\and
Shapley Additive Explanations
}
\end{abstract}

\section{Context and Objectives}

Ensemble learning has emerged as a powerful paradigm in machine learning, leveraging the diversity of multiple base models to achieve improved generalization and robustness~\cite{CAMPAGNER2023241}. Among ensemble methods, stacked generalization, or \textit{stacking}, stands out for its ability to integrate heterogeneous models by training a meta-learner on the predictions of base models, thus capturing complementary patterns in the data~\cite{CAMPAGNER2023241,xstacking}. In this two-stage architecture, a set of $K$ base learners $f_1, \dots, f_K$ are trained on a dataset $\mathcal{D} = \{(\mathbf{x}_i, y_i)\}_{i=1}^m$, where \( \mathbf{x}_i \in \mathbb{R}^d \) represents a d-dimensional vector of characteristics, and \( y_i \in \mathbb{R} \) (or \( y_i \in \{0,1\} \) for binary classification) denotes the target variable. Each model produces an individual output prediction such as $\hat{y}_{i}^{(k)} = f_k(\mathbf{x}_i)$, \ot{$k \in [1,...,K]$}. These outputs form a new dataset \( \mathcal{D}' = \{(\hat{\mathbf{y}}_i, y_i)\}_{i=1}^{m} \), where \ot{\( \hat{\mathbf{y}}_i = (\hat{y}_i^{(1)}, \dots, \hat{y}_i^{(K)}) \)}. The meta-learner $\mathcal{E}$ is then trained on \( \mathcal{D}' \) to map the new learning space \( \hat{\mathbf{y}}_i = (\hat{y}_i^{(1)}, \dots, \hat{y}_i^{(K)}) \) to the final output of the ensemble $\hat{y}_i = \mathcal{E}(\hat{\mathbf{y}}_i)$.

Stacking has proven highly effective in a variety of machine learning tasks due to its ability to combine different base models and identify intricate decision boundaries~\cite{CAMPAGNER2023241}. However, despite its success, stacking suffers from two critical limitations. 
First, the meta-learner input space—formed by the predictions of base models $\hat{\mathbf{y}}_i$— is often weakly informative~\cite{xstacking}. If base models are highly correlated or do not capture complementary aspects of the data, their aggregated predictions can offer a low diversity feature space or limited value, reducing the meta-learner’s ability to discriminate between different classes or response levels. 
Second, stacking models are inherently difficult to interpret. The meta-learner aggregates the outputs of multiple base models, many of which may themselves be opaque, without exposing the rationale behind the final decision. This nested black-box structure complicates feature attribution and makes it challenging to explain how individual input variables influence the ensemble's predictions. In domains such as healthcare or finance, this lack of transparency raises concerns about fairness and user trust.


To overcome these issues, we introduce \textit{XStacking}, an inherently explainable stacked ensemble framework. XStacking enhances the meta-learning space by integrating model-agnostic explanations—specifically, Shapley values~\cite{shap} as additional input features. For each instance $\mathbf{x}_i$, the second-stage input includes a concatenation of feature importance vectors $\phi_k(f_k(\mathbf{x}_i))$, providing a richer and more discriminative representation. These vectors are concatenated across models to form an enriched representation $\hat{\mathbf{y}}_i^{*} = (\phi_1(f_1(\mathbf{x}_i)), \dots, \phi_K(f_K(\mathbf{x}_i)))$, used as input to the meta-learner $\mathcal{E}$. This augmentation improves data separability, enables feature-level interpretability, and mitigates redundancy among base 




\vspace{-0.1cm}
\section{Proposed Approach}
\label{sec:proposed_approach}

XStacking is an explanation-guided ensemble method that builds a meta-learner using not only the predictions of base models but also their explanations. It is inspired by human decision-making, which integrates insights from multiple individuals by considering both their decisions and the rationales behind them. This reasoning is modeled using feature importance scores—specifically, Shapley values—produced by the base classifiers. These explanations are concatenated to form an enriched learning space that guides the decision-making process of the ensemble stacking classifier.

Given training data  $\mathcal{D} = \{(\mathbf{x}_i, y_i)\}_{i=1}^m$ ($\mathbf{x}_i \in \mathbb{R}^d, y_i \in \mathcal{Y}$) with $m$ instances, \( d \)-dimensional feature vector, and their class labels $\mathcal{Y}$, we train $k$ base learners $F = \{f_1, f_2, \ldots, f_K\}$ on $\mathcal{D}$ using $K$-fold cross-validation. For each learner $f_k$, we compute Shapley values $\phi_k$ using SHAP~\cite{shap}, representing the feature importance scores for each instance $\mathbf{x}_i$ with respect to its prediction $\hat{y}_{i}^{(k)} = f_k(\mathbf{x}_i)$.
The resulting Shapley vectors $\Phi = [\phi_1, \phi_2, \dots, \phi_K]$ are concatenated to form an enriched feature representation for each instance. This results in a transformed dataset $\mathcal{D}'$, where each instance includes $k \times d$ explanation-based features. The meta-learner $\mathcal{E}$ is then trained in $\mathcal{D}'$ in the second stage of learning. The general procedure of XStacking is detailed in Algorithm~\ref{alg:xstacking}, which involves three main steps.

\begin{algorithm}[H]
\caption{XStacking algorithm.}
\label{alg:xstacking}
\KwIn{Training data $\mathcal{D} = \{(\mathbf{x}_i, y_i)\}_{i=1}^m$ ($\mathbf{x}_i \in \mathbb{R}^d, y_i \in \mathcal{Y}$)}
\KwOut{An ensemble classifier $\mathcal{E}$}

\textbf{Step 1:} Learn first-level classifiers\\
\For{$j \gets 1$ \textbf{to} K}{
    Learn a base classifier $f_j$ based on $\mathcal{D}$\\
}

\textbf{Step 2:} Construct the new learning space\\
\For{$j \gets 1$ \textbf{to} K}{
    $\phi_j$ =\texttt{shapley\_values}$(f_j(\mathcal{D}))$\\
    
\For{$i \gets 1$ \textbf{to} $m$}{

    Construct the explanations set $\{\hat{\mathbf{y}}_i, y_i\}$, where $\hat{\mathbf{y}}_i = \{\phi_1(f_1(\mathbf{x}_i)), \phi_2(f_2(\mathbf{x}_i)), \dots, \phi_K(f_K(\mathbf{x}_i))\}$\\
}}

\textbf{Step 3:} Train the meta-model classifier\\
    Learn a new classifier $f^*$ based on the newly constructed dataset $\mathcal{D'} = \{(\hat{\mathbf{y}}_i, y_i)\}_{i=1}^m$ ($\hat{\mathbf{y}}_i \in \mathbb{R}^{k \times d}, y_i \in \mathcal{Y}$)\\

\Return $\mathcal{E}(\texttt{x}) = f^*\{\phi_1(f_1(\texttt{x})), \phi_2(f_2(\texttt{x})), \dots, \phi_K(f_K(\texttt{x}))\}$\\

\end{algorithm}

\begin{description}
    \item[Step 1:] Learn first-level classifiers based on the original training dataset. We can apply different classification methods and/or sampling methods to generate base classifiers (heterogeneous base classifiers).

    \item[Step 2:] Construct a new learning space based on the Shapley values from the output of base classifiers $f_k$. Here, the Shapley values $\phi_k$ from output predicted labels of the first-level classifiers are regarded as new features, and the original class labels are kept as the labels in the new dataset. For instance, for each sample $\{\mathbf{x}_i,y_i\} \in \mathcal{D}$, we construct a corresponding set $\{\hat{\mathbf{y}}_i, y_i\}$  in the new learning space $\mathcal{D'}$, where $\hat{\mathbf{y}}_i~=~\{\phi_1(f_1(\mathbf{x}_i)), \phi_2(f_2(\mathbf{x}_i)), \dots, \phi_K(f_K(\mathbf{x}_i))\}$.

    \item[Step 3:] Learn a second-level meta-classifier based on the newly constructed base of explanations. 
\end{description}

Once the second-level classifier is generated, it can be used to combine the first-level classifiers. For an unseen example $\texttt{x}$, its predicted class label of stacking is $f^*\{\phi_1(f_1(\texttt{x})), \phi_2(f_2(\texttt{x})), \dots, \phi_K(f_K(\texttt{x}))\}$, where $\{f_1, f_2,\dots, f_K\}$ are first-level classifiers, $\{\phi_1, \phi_2,\dots, \phi_K\}$ are the shapley values of first-level classifiers 
and $f^*$ is the second-level meta-learner.

\section{Experimental study}

We evaluate XStacking across 29 classification and regression datasets, focusing on three key questions: \textbf{(RQ1) Effectiveness}, \textbf{(RQ2) Efficiency}, and \textbf{(RQ3) Explainability}. Performance was compared with traditional stacking using SVM and XGBoost meta-learners. Statistical significance was tested using Wilcoxon signed-rank tests.
The base layer includes diverse model families: tree-based models (decision trees), linear models (linear and logistic regression), and neural networks (multilayer perceptrons) as first-stage base learners for both classification and regression tasks.


\textbf{RQ1—Effectiveness.}  
We compared XStacking with traditional stacking using both SVM and XGBoost as meta-learners. In classification, XStacking achieved equal or better accuracy on 16 of 17 datasets with the SVM meta-learner and on 14 of 17 datasets with XGBoost. Predictive precision improvements were notable on datasets like \textit{adult} (+2.6\%) and \textit{vehicle} (+5.9\%). In regression tasks, XStacking outperformed traditional stacking in 11 out of 12 datasets using the SVM meta-learner. For example, on \textit{cpu\_small}, MSE dropped from 22.4 to 11.3 (SVM), and further to 7.6 (XGBoost). Across all tasks, the improvements were statistically significant under a Wilcoxon signed-rank test ($p < 0.01$), confirming the robustness of the results.

\textbf{RQ2—Efficiency.}  
The integration of SHAP values introduces moderate computational overhead, primarily during first-stage explanation extraction. However, the cost is offset by performance gains and remains acceptable for most datasets. The meta-learner training time remains comparable to standard 

\textbf{RQ3—Explainability.}  
Unlike traditional black-box stacking, XStacking integrates SHAP values into the second-stage learning process, making the decision pipeline inherently interpretable. This enables the second-stage model to learn from feature attributions, producing inherently interpretable outputs. Qualitative analysis shows that important features identified by XStacking align with domain expectations and enhance user trust in the model's decisions.

\noindent\textbf{Summary.}
XStacking consistently delivers improved performance across diverse datasets while offering built-in interpretability. It provides a principled way to integrate explanation signals into ensemble learning and demonstrates the potential of explanation-guided modeling. The detailed results and their in-depth analysis are presented in the original publication\,\cite{xstacking}. The source code is available on Github\,: \url{https://github.com/LeMGarouani/XStacking}.

\vspace{-0.3cm}
\section{Conclusion}
\vspace{-0.27cm}
XStacking demonstrates that explainability and accuracy can be combined in ensemble models. By incorporating SHAP values in second-stage learning, we achieve a stacking framework that is powerful and interpretable.
Future work includes applying the method to time series data, extending to other explanation techniques (e.g., LIME, gradient-based attribution methods), and investigating the efficiency of XStacking in multimodal ensemble learning, where multiple data modalities are combined. This presents an exciting direction. This would involve developing methods to handle diverse explanation types across different modalities while ensuring interpretability across heterogeneous data sources.

\vspace{-0.35cm}
\bibliographystyle{splncs04}
\bibliography{bibliography}
\vspace{-0.5cm}

\end{document}